\documentclass{article}

\usepackage{arxiv}

\usepackage[utf8]{inputenc} 
\usepackage[T1]{fontenc}    
\usepackage{hyperref}       
\usepackage{url}            
\usepackage{booktabs}       
\usepackage{amsfonts}       
\usepackage{nicefrac}       
\usepackage{microtype}      
\usepackage{lipsum}
\usepackage{graphicx}
\graphicspath{ {./images/} }

\usepackage{graphicx}
\usepackage{booktabs}
\usepackage{amssymb}
\usepackage{amsmath}
\usepackage{subcaption}
\usepackage{url}

\title{A Neurorobotics Approach to Behaviour Selection based on Human Activity Recognition}

\author{
 Caetano M. Ranieri \\
 Institute of Mathematical and Computer Sciences \\
 University of Sao Paulo \\
 Sao Carlos, SP, Brazil \\
  \texttt{cmranieri@alumni.usp.br} \\
\And
  Renan C. Moioli \\
  Digital Metropolis Institute\\
  Federal University of Rio Grande do Norte\\
  Natal, RN, Brazil \\
  \texttt{renan.moioli@imd.ufrn.br}
\And
 Patricia A. Vargas \\
  Edinburgh Centre for Robotics\\
  Heriot-Watt University\\
  Edinburgh, Scotland, UK \\
  \texttt{p.a.vargas@hw.ac.uk} \\
\And
 Roseli A. F. Romero \\
 Institute of Mathematical and Computer Sciences \\
 University of Sao Paulo \\
 Sao Carlos, SP, Brazil \\
  \texttt{rafrance@icmc.usp.br} \\
}

\begin{document}
\maketitle
\begin{abstract}
Behaviour selection has been an active research topic for robotics, in particular in the field of human-robot interaction. For a robot to interact effectively and autonomously with humans, the coupling between techniques for human activity recognition, based on sensing information, and robot behaviour selection, based on decision-making mechanisms, is of paramount importance. However, most approaches to date consist of deterministic associations between the recognised activities and the robot behaviours, neglecting the uncertainty inherent to sequential predictions in real-time applications. In  this  paper,  we  address  this  gap  by  presenting  a  neurorobotics  approach based  on  computational  models  that  resemble  neurophysiological  aspects  of living beings. This neurorobotics approach was compared to a non-bioinspired, heuristics-based approach. To evaluate both approaches, a robot simulation is developed, in which a mobile robot has to accomplish tasks according to the activity being performed by the inhabitant of an intelligent home. The outcomes of each approach were evaluated according to the number of correct outcomes provided by the robot. Results revealed that the neurorobotics approach is advantageous, especially considering the computational models based on more complex animals.
\end{abstract}

\keywords{Behaviour selection \and Human activity recognition \and Robot simulation \and Neurorobotics \and Bioinspired computational model}

\section{Introduction}
\label{sec:introduction}

Truly autonomous behaviour is still not the norm for robots designed to interact socially with humans \cite{Clabaugh2019EscapingRobotics}. In general, behaviour selection has been an active research topic for robotics in general, and human-robot interaction in particular \cite{Ko2018BehaviorThought}.
In this context, the need for a real-time understanding of human actions is of paramount importance for the robotic agent to behave proactively and effectively. Such a requirement could be achieved with techniques for human activity recognition \cite{Mojarad2018HybridRobots}.

When dealing with complex modalities (e.g., videos or data from inertial units), activity recognition approaches often rely on machine learning.
For instance, video-based activity recognition have been approached by architectures based on convolutional and recurrent neural networks \cite{Herath2017GoingSurvey,Ma2019TS-LSTMRecognition}.
For inertial data, similar architectures have been proposed, processing either raw data \cite{Ordonez2016DeepRecognition,Garcia2019TemporalSensors} or descriptors obtained through feature extraction methods \cite{StevenEyobu2018FeatureNetwork,Ashry2020CHARM-Deep:Smartwatch}.
To provide a wider range of possibilities, robots may act symbiotically with other pervasive devices, such as wearable technologies or ambient sensors in intelligent environments, which may provide additional capabilities for sensing and acting based on application-specific components~\cite{Bacciu2019AnEnvironments}.
When synchronised data from different sensors are available, activity recognition techniques may rely on multiple sensor modalities to provide more accurate results, giving rise to techniques for multimodal activity recognition \cite{Lu2019AutonomousSensors,Imran2020EvaluatingRecognition,Ranieri2020UncoveringApproach}.

Although human activity recognition has been a quite fertile field of research, few approaches have been developed to link the outputs from those algorithms into actual response behaviours from a robot.
Related works usually consist of direct associations between the recognised activities and the response behaviours \cite{Georgievski2017PlanningBuildings,Li2019Real-TimeMechanism,RodriguezLera2020AScenarios}.
One of the possibilities consist of combining computational neuroscience to the robotics scenarios, characterising the field of neurorobotics~\cite{VanDerSmagt2016Neurorobotics:Action}, which may build upon different biological aspects that influence the behaviour of living beings.

Li \textit{et al.} \cite{Li2019CombinedSurvey} provided a comprehensive survey on neurorobotics systems (NRS) and the different components that may integrate them.
According to the authors, a generalised framework can be depicted for most NRSs in the literature, composed of a \textit{simulated brain}, which is fed with sensory signals from a \textit{body} and turns them into control signals for a \textit{hierarchical controller}, responsible for decoding these signals into control commands for the body, which actuates and senses an external environment.
Bioinspired strategies may be introduced to different aspects of the framework, according to the required task of a particular study.

The basal ganglia, a group of subcortical nuclei present in the vertebrate's brain, is known to have an important role in action selection mechanisms, especially regarding striatal circuits \cite{Markowitz2018TheSelection}.
The so-called direct and indirect pathways are characterised by competitive or complementary functions that mediate the excitation of the motor system based on inputs from the motivational system of an individual, deciding whether to ''go'' or to ''stop'' performing a certain behaviour \cite{Bariselli2019ASelection}.
The potential roles of such a mechanism in robotic frameworks have also been evaluated, including simulations in which bioinspired networks receiving different stimuli are expected to respond with different behaviours, resulting in cooperative interactions that produce robot behaviours \cite{Bahuguna2018ExploringFramework}.


In this paper, we present a neurorobotics model which embeds computational models of the basal ganglia-thalamus-cortex (BG-T-C) circuit \cite{Kumaravelu2016ADisease,ranieri2021towardsPD} to provide a decision-making mechanism for a robot - in this context, we may call it a \textit{neurorobot}. The neurorobotics approach has been proposed for enhancing the decision-making mechanism, as suggested by related researches in neurorobotics \cite{Liang2019APrediction,Mulcahy2020BasalStates}.
It consisted of simulating neurophysiological aspects within a cognitive framework, in which different stimuli was introduced to certain brain structures within the circuit, according to real-time outputs of the activity recognition module. The resulting spike trains from the neurorobotics model were then converted to neural firing frequencies across brain regions, which would be further decoded using convolutional neural networks, in order to infer the most suitable response behaviour for the robot.

The application scenario is a simulated smart home, in which an activity recognition model, presented in \cite{Ranieri2021ActivitySensors} for the HWU-USP activities dataset \cite{Ranieri2021HumanSensors}, was employed in human-robot interaction tasks, using a mobile robot.
In summary, the robot needs to produce response behaviours according to the contextual information inferred by the user (i.e., the recognised activity).


The neurorobotics approach, which is the central contribution of this work, was compared to a heuristics approach, in which a deterministic behaviour selection mechanism was considered using simple heuristics that associate recognised activities to robot behaviours.
This neurorobotics approach embedded two computational models, one that resembled neurophysiological data of rodents (i.e., the rat model), and one that resembled data from marmoset monkeys (i.e., the primate model). 

The different factors considered for this study were evaluated according to the relative number of correct outcomes of the robot simulation. Considering the activity recognition framework, the results have confirmed that more accurate classifiers for the activity recognition module led to a greater number of robot tasks successfully completed.
Although the performances of the heuristic and neurorobotics approaches varied according to the computational model embedded, the study confirmed that the most complex neurorobotics model (i.e., the marmoset-based model of the BG-T-C circuit) led to an increased performance in relation to the heuristic approaches when a more accurate activity recogniser was considered (i.e., the video-based classifier).

The remainder of this paper is organised as follows.
The brain structures considered for this neurorobotics approach and the computational modelling adopted are presented in Section \ref{sec:motor-loop}.
In Section \ref{sec:proposed-system}, are presented the general aspects of the robotic system, and the integration between each of its modules.
In Section \ref{sec:neuromodel}, the neurorobotics approach is detailed.
In Section \ref{sec:methods}, the methods and implementations are depicted.
The corresponding results are presented in Section \ref{sec:results} and discussed in Section \ref{sec:discussion}.
Finally, the concluding remarks and directions for future research are provided in Section \ref{sec:conclusion}.

\section{The BG-T-C Circuit and Original Computational Models}
\label{sec:motor-loop}
\begin{figure}[!ht]
    \centering
    \begin{subfigure}{0.45\textwidth}
        \centering
        \includegraphics[scale=0.6]{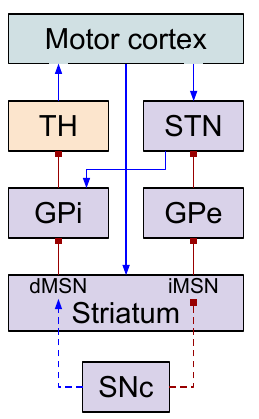}
        \caption{Classic model, as described by \cite{McGregor2019CircuitDisease}.}
        \label{sfig:classic-model}
    \end{subfigure}
    \begin{subfigure}{0.45\textwidth}
        \centering
        \includegraphics[scale=0.6]{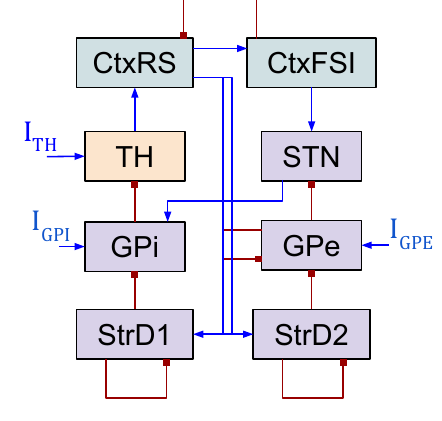}
        \caption{Computational model, as designed by \cite{Kumaravelu2016ADisease} for rodent data, and adapted by \cite{ranieri2021towardsPD} for primate data.}
        \label{sfig:kumaravelu}
    \end{subfigure}
    \caption{Schematic representations of the classic and computational models of the BG-T-C circuit.
    In the connections, excitatory synapses are shown as blue arrows, and inhibitory synapses, as red squares.}
    \label{fig:bg_circuit}
\end{figure}

In this section, we present the basic concepts on the brain structures present in the basal ganglia-thalamus-cortex (BG-T-C) circuit, and the original computational modelling. The BG-T-C circuit, illustrated in Figure \ref{fig:bg_circuit}, is formed by the \textit{motor cortex} (M1), the \textit{thalamus} (TH), and the \textit{basal ganglia} (BG), the latter composed of a subset of structures: the \textit{striatum} (Str), the \textit{globus pallidus}, divided into \textit{pars interna} (GPi) and \textit{pars externa} (GPe), the \textit{subthalamic nucleus} (STN), and the \textit{substantia nigra}, divided into \textit{pars compacta} (SNc) and \textit{pars reticulata} (SNr).

In \cite{McGregor2019CircuitDisease}, is provided a discussion about the mechanisms of this circuit and presented models to describe it. The most useful model to explain the connections within this circuit, especially those affected by PD, is the so-called classic model, illustrated in Figure~\ref{sfig:classic-model}.

The pathways start with an excitatory connection from the cortex to the striatum, which projects its output neurons, named \textit{medium spiny neurons} (MSN), to other structures inside the BG.
In the direct pathway, the direct MSN (dMSN) inhibits the GPi, which reduces its inhibition to the TH. Then, it excites the motor cortex. In the indirect pathway, the indirect MSN (iMSN) inhibits the GPe, which reduces its inhibition to the STN, which excites the GPi.
Thus, this results on inhibition of the TH and absence of excitatory outputs to the motor cortex.
In other words, the direct pathway excites the cortex (i.e., positive feedback loop), while the indirect pathway inhibits it (i.e., negative feedback loop).

In \cite{Kumaravelu2016ADisease}, a computational model of the BG-C-T circuit, originaly developed to study the underlying mechanisms of Parkinson's Disease (PD), was proposed and implemented based on neural data from healthy and PD-induced (i.e., 6-OHDA lesioned) rats \cite{Kita2011CorticalGanglia}.
Eight brain structures were modelled and connected based on a simplified version of the classic model (see Figure \ref{sfig:kumaravelu}).
In particular, the direct and indirect pathways were modelled separately representing the MSN modulation by D1 and D2 dopamine receptors in the striatum (i.e., StrD1 and StrD2, respectively).
The cortex is represented by regular spiking (RS) excitatory neurons and fast spiking (FSI) inhibitory interneurons (i.e., CtxRS and CtxFSI, respectively).
A bias current was added in the TH, GPe, and GPi, accounting for the inputs not explicitly modelled. 
This model was designed with the ability to shift from the simulation of healthy to the PD status, which is done by altering certain conductances.

Although all mammals have a similar set of BG structures that are similarly connected with thalamic and cortical structures, subtle differences between species may be found, with primates being more similar to humans than rodents \cite{Lienard2014,Koprich2017AnimalDevelopment,Dawson2018}.
A data-driven approach was proposed in \cite{ranieri2021towardsPD} to obtain a primate-based computational model of the BG-T-C circuit and the mechanisms of PD.
The resulting marmoset computational model was evaluated based on the differences between healthy and PD individuals, with respect to the spectral signature of the brain activity \cite{Tinkhauser2017BetaMedication}, the dynamics of the firing rates of neurons across brain regions \cite{VanAlbada2009}, and the coherence between spike trains \cite{Halje2019}.

The implementation used in \cite{ranieri2021towardsPD} built on a Python translation of the original computational model of \cite{Kumaravelu2016ADisease}, originally made by \cite{Romano2020EvaluationDisease} using the NetPyNE framework and the libraries from the NEURON simulator \cite{Dura2019}.
Based on the results of the machine learning framework, a practical setup of either the rat or marmoset computational models was made available. The adaptations performed in this work to the original computational models (see Subsection~\ref{ssec:computational-model}) were based on the code made available by the authors. For all neurorobotics model evaluations, we considered both the rat and primate computational models, always with the healthy state set on.



\section{Integrated System}
\label{sec:proposed-system}

\begin{figure}
    \centering
    \begin{subfigure}{\textwidth}
        \centering
        \includegraphics[scale=0.65]{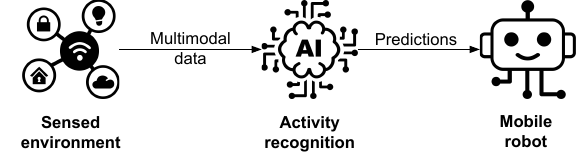}
        \caption{Heuristics approach: the predictions from the activity recognition module are fed directly to the mobile robot.}
        \label{sfig:heuristics-approach}
    \end{subfigure}
    \begin{subfigure}{\textwidth}
        \centering
        \includegraphics[scale=0.65]{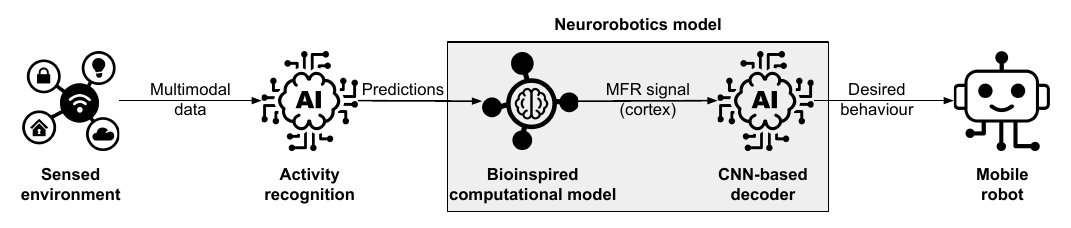}
        \caption{Neurorobotics approach: the predictions are used as stimuli for the embedded, bioinspired computational model of the BG-T-C circuit, which simulates neural activity that is further interpreted by a CNN-based decoder, responsible for deciding the behaviour to be performed by the robot. Both the bioinspired computational model and the CNN-based decoder compose the neurorobotics model presented in this research.}
        \label{sfig:neurorobotics-approach}
    \end{subfigure}
    \caption{Interaction between modules for the application scenario proposed.}
    \label{fig:pipeline}
\end{figure}

The modules of the application scenario, and the interactions between them, are illustrated in Figure~\ref{fig:pipeline}. In this scenario, the human activities are inferred by a machine learning algorithm, and the supporting behaviours are performed by a mobile robot placed in a simulated environment, composing an ambient assisted living (AAL) application~\cite{calvaresi2017exploring}.

The general information flow was: given the multimodal data provided by a set of sensors within a \textit{sensed environment}, apply an \textit{activity recognition module} to classify such data into a set of predefined human activities, and produce correspondent response behaviours for a \textit{mobile robot}.

The neurorobotics approach was compared to a heuristics approach. The heuristics approach (Figure~\ref{sfig:heuristics-approach}) consisted of associating the predictions of the activity recognition module to response behaviours based on simple heuristics, presented  in Subsection~\ref{ssec:robot-simulation}. In the neurorobotics approach (Figure~\ref{sfig:neurorobotics-approach}), the predictions from the activity recognition module were employed to stimulate a \textit{bioinspired computational model}, whose outputs (i.e., neural firing frequencies of brain simulated regions) were decoded by a \textit{CNN-based decoder}, which provided the decisions for the mobile robot.


More specifically, a \textit{sensed environment} consisted of a previously collected dataset \cite{Ranieri2021HumanSensors}, composed by a set of recording sessions $X$, with each $x \in X$ associated to an activity $a \in A = \{a_1,\dots,a_{N_A}\}$, where $N_A$ is the number of classes  (i.e., labels) considered for this dataset.
The function describing these associations is given by Equation~\ref{eq:fa}.

\begin{equation}
    \label{eq:fa}
    \begin{split}
        f_A: X \rightarrow A \Longleftrightarrow f_A(x)=a,\\
        x \in X, \quad a \in A
    \end{split}
\end{equation}

Each data tuple $x(t)$ comprises a segment, with a previously defined length, of a recording session $x$ starting at timestep $t \in T_x = \{1,\dots,N_{T_X}\}$, equally spaced among them, to be segmented from $x$.
The \textit{activity recognition} module is a machine learning classifier $g \in G$, which might associate a recording session $x \in X$ at timestep $t \in T_X$ to an activity $a \in A$, through a prediction vector $y_x^t \in Y$ (see Equation \ref{eq:g}).
In other words, considering that $a$ is unknown at inference time, the inference model $g$, learned from labelled samples, provides a prediction vector $y_x^t$, where $y_x^t(a)$ is the probability that a given input $x(t)$ corresponds to activity $a$.

\begin{equation}
    \label{eq:g}
    \begin{split}
        g: X \times T \rightarrow Y \Longleftrightarrow g(x, t)=y_x^t,\\
        x \in X, \quad t \in T, \quad y_x^t \in Y
    \end{split}
\end{equation}

The application scenario was designed so that each activity in $A$ was associated to a desired response for the mobile robot.
We defined a set of response behaviours $B= \{b1,\dots,b_{N_B}\}$, so that each human activity $a \in A$ can be, but not necessarily is, associated to a response behaviour of the robot.
The ''no action'' behaviour is denoted as $b_\emptyset$.
Hence, the function that associates recognised activities to response behaviours is given by Equation~\ref{eq:fb}.

\begin{equation}
    \label{eq:fb}
    \begin{split}
            f_B: A \rightarrow B \cup \{b_\emptyset\} \Longleftrightarrow f_B(a)=b,\\
            a \in A, \quad b \in B \cup \{b_\emptyset\}
    \end{split}
\end{equation}

The robot simulation would be considered successfully completed if:

\begin{itemize}
    \item For an activity $a$ being performed in the environment in a session $x$, the robot completed an expected response behaviour $b \in B$ before $x$ was finished; or
    \item No response behaviour was expected (i.e., $f_B(a)=b_\emptyset$) and the robot did not complete any of the behaviours in $B$.
\end{itemize}

It is worth to notice that, according to this evaluation policy, besides an accuracy requirement (i.e., the correct behaviour must be given in response to a human activity), there was also a time constraint that must be satisfied (i.e., if required, the response behaviour must be completed while the human is still performing the given activity).

Since, by definition, $f_A(x)=a$ is not known at runtime, and can only be inferred by a classifier $g \in G$ as successive prediction vectors $y_x^t$ are provided, a decision-making mechanism was needed to perform adaptive decisions based on partial, time-localised predictions.
To this aim, we proposed the neurorobotics model presented in Section \ref{sec:neuromodel}, and compared it to a simple heuristics-based approach as described in Section \ref{sec:methods}.

\section{The Neurorobotics Model}
\label{sec:neuromodel}

The neurorobotics model embeds the bioinspired computational model and the CNN-based decoder (see Figure \ref{fig:pipeline}).
It consists of simulating and decoding the neurophysiological mechanisms of the basal ganglia-thalamus-cortex (BG-T-C) circuit in mammals (see Section \ref{sec:motor-loop}), responsible for abilities such as motor control, decision-making, and learning \cite{Girard2008WhereSelection,Liang2019APrediction,Mulcahy2020BasalStates}.
As already stated in Section~\ref{sec:motor-loop}, both the rat-based \cite{Kumaravelu2016ADisease} and the marmoset-based \cite{ranieri2021towardsPD} computational models were evaluated as a decision-making mechanism of the neurorobotics model.

\subsection{Bioinspired Computational Model}
\label{ssec:computational-model}



\begin{figure}[!ht]
    \centering
    \begin{subfigure}{0.45\textwidth}
        \centering
        \includegraphics[scale=0.6]{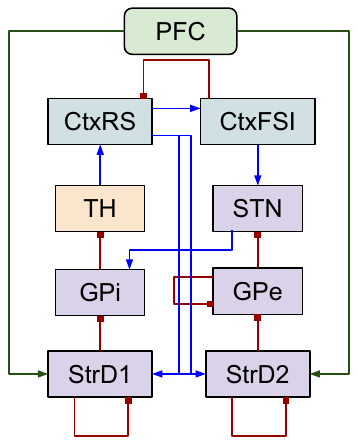}
        \caption{Schematic representation of the computational model as adapted in this work.
    In the connections, excitatory synapses are shown as blue or green arrows, and inhibitory synapses, as red squares. The blue arrows and red squares correspond to the original synapses as designed by \cite{Kumaravelu2016ADisease} and adapted by \cite{ranieri2021towardsPD}, while the green arrows are the adaptations provided in this work to allow the stimulation of the circuit in the context of the application scenario proposed.}
        \label{sfig:adapted-model}
    \end{subfigure}
    \begin{subfigure}{0.45\textwidth}
        \centering
        \includegraphics[scale=0.6]{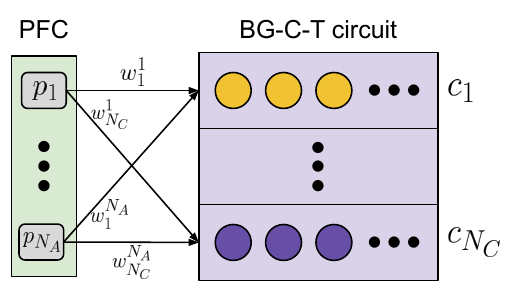}
        \caption{Predictions from the activity recognition module are interpreted as stimulation originated on the prefrontal cortex (PFC), which selectively stimulates different populations of the computational model, each associated to one response behaviour of the robot.}
        \label{sfig:adapted-connections}
    \end{subfigure}
    \caption{Adapted version of the computational model of the BG-T-C circuit.}
    \label{fig:cm_adapted}
\end{figure}

Motivated by the work of \cite{Mulcahy2020BasalStates}, two key modifications were introduced to the computational models of the BG-T-C circuit adopted in this work \cite{Kumaravelu2016ADisease,ranieri2021towardsPD}.
First, an additional structure, called prefrontal cortex (PFC), was included as a variable source of excitatory stimuli towards the striatum (see Figure~\ref{sfig:adapted-model}).
Second, $N_C=N_B$ populations of neurons were implemented as independent channels $c \in C$, each associated to exactly one response behaviour $b \in B$ (see Figure~\ref{sfig:adapted-connections}), as defined in Equation~\ref{eq:fc}.

\begin{equation}
    \label{eq:fc}
    \begin{split}
        f_C: B \rightarrow C \Longleftrightarrow f_C(c)=b,\\
        b \in B, \quad c \in C
    \end{split}
\end{equation}

At each timestep, the channels $c \in C$ received a stimulus $s \in S = \{s_1,\dots,s_{N_C}\}$, whose intensity was based on the linear combination between a prediction vector $y_x^t$ and a weight function $f_W$, given by Equation~\ref{eq:weights}.
The actual value of $s$ is given by function $f_S$, defined as in Equation~\ref{eq:stimulus}.

\begin{equation}
    \label{eq:weights}
    \begin{split}
        f_W: C \times A \rightarrow \{0,1\} \Longleftrightarrow f_W(c, a) =
        \begin{cases}
        1, \quad \text{if } f_B(a) = f_C(c)\\
        0, \quad \text{otherwise}
        \end{cases}\\
        c \in C, \quad a \in A
    \end{split}
\end{equation}

\begin{equation}
    \label{eq:stimulus}
    \begin{split}
        f_S: C \times A \rightarrow S \Longleftrightarrow f_S(c, a) = s = \sum_{a \in A} f_W(c,a) \cdot y_x^t(a)\\
        c \in C, \quad a \in A, \quad s \in S
    \end{split}
\end{equation}

Considering that, as ensured by the softmax activation on the classifiers, $\sum_{a \in A} y_x^t(a) = 1$, then $s \in [0,1] \mu A / cm^2$, which has shown to be a stable, biologically plausible interval.
For a recording sequence $x$, the set of prediction vectors $y_x$ is employed to  update periodically each stimulus $s \in S$, during the course of a corresponding simulation of the computational model (not to be confused with the robot simulation).
For each simulation, $N_{T_\text{sim}}$ subsequent updates would be done for all $s \in S$, computed for the timesteps in $y_x$.

A simulation, after finished, produced a spike train for each of the brain regions modelled, contemplating all its length (i.e., all $N_{T_\text{sim}}$ updates were considered). The neural firing frequencies were computed according to \cite{Lansky2004MeanRate}, with the parameters detailed in Subsection \ref{ssec:neurorobotics-configurations}, and summed across each region of each channel, resulting in $N_R \cdot N_C$ output signals for each simulation, each with length $L_U$, where $N_R = 8$ is the number of regions (see Figure \ref{fig:bg_circuit}b).

Formally, let $u_x^{g,m} \in U$ be defined as the output for a given simulation, where $x \in X$ is a recording session, $g \in G$ is the classifier employed for activity recognition, and $m \in M$, a computational model.
Therefore, let a simulation be defined as $f_U$ (see Equation \ref{eq:fu}), whose output is as a multivariate time-series with $N_R \cdot N_C$ variables and $L_U$ timesteps.

\begin{equation}
    \label{eq:fu}
    \begin{split}
        f_U: X \times G \times M \rightarrow U \Longleftrightarrow f_U(x, g, m) = u_x^{g,m}\\
        (x, g, m) \in X \times G \times M, \quad u \in U
    \end{split}
\end{equation}

After the simulations were completed, the spike trains at the cortex populations were converted into temporal signals (i.e., neural firing frequencies) based on the mean firing rates across brain regions \cite{Lansky2004MeanRate}. The resulting signals were segmented in smaller windows and applied to train and evaluate a convolutional neural network (CNN), which would be employed to determine the decision of the robot at each timestep of the robot simulation (i.e., the \textit{CNN-based decoder}). More details on the implementation of the CNN-based decoder are presented in the next section.

\subsection{CNN-Based Decoder}
\label{ssec:cnn-decoder}

Each simulation of the computational model provided the summed neural firing frequencies of each channel and brain region, generating a data structure $u_x^{g,m}$, associated to the whole recording session that generated it.
As a requirement to provide a realistic scenario for the robot simulation, time-localised decisions were required, which must be taken based only in past events.
In other words, at a timestep $t_\text{robot}=i$ of the robot simulation, only predictions obtained on timesteps $t_j, j<i$ could be taken into account when providing a response behaviour to the robot.

To fulfil this requirement, each instance $u_x^{g,m}$, correspondent to the recording session $x \in X$ in the set of conditions $g \in G$ and $m \in M$ (see Equation \ref{eq:fu}), was segmented in windows of $N_V$ timesteps, with partial superposition, producing $N_\text{segs}$ segments.
Considering $N_X$ recording sessions in a given set of conditions, the function $f_v$ would generate a total of $N_X \cdot N_\text{segs}$ instances $v \in V$, as defined in Equation \ref{eq:f_v}.

\begin{equation}
    \label{eq:f_v}
    \begin{split}
        f_v: U \times T \rightarrow V \Longleftrightarrow f_v(u_x^{g,m}, t) = v = u_x^{g,m}[t,t+N_V]\\
        u_x^{g,m} \in U, \qquad t \in T \quad | \quad t+N_V<L_U, \qquad v \in V
    \end{split}
\end{equation}

The resulting segments were employed to train a machine learning decoder (i.e., the CNN-based decoder).
We considered only the cortex regions to compose the input tuples for the decoder, aiming to preserve biological plausibility regarding this aspect.
Given that each channel of the computational model has two cortex regions (i.e., cortex RS and FSI), and that the experiments were performed with $N_C$ channels, associated to the response behaviours $b \in B$, the resulting instances $v$ had shape $N_V \times 2 N_C$.
The decoder $f_Q$ might be trained to provide a decision vector $q_x^t$, which corresponds to the probability that a given segment of cortex firing frequencies, given by$v = f_v(u_x^{g,m},t)$, might be associated to a behaviour in $B \cup \{b_\emptyset\}$.
This decoding function may be defined as in Equation \ref{eq:q}.

\begin{equation}
    \label{eq:q}
    \begin{split}
        f_Q: V \rightarrow Q \Longleftrightarrow f_Q(v)=q_x^t,\\
        v \in V, \quad q_x^t \in Q 
    \end{split}
\end{equation}

We have adopted a one-dimensional convolutional neural network (CNN) as decoder, which has shown to provide state-of-the-art results in related work~\cite{Ranieri2020UnveilingNetworks} (for the architectural choices and hyperparameter settings, see Subsection \ref{ssec:neurorobotics-configurations}).
Classification metrics were provided considering that the categorical output is chosen according to Equation \ref{eq:d_q}, where $d_Q$ corresponds to a response behaviour.

\begin{equation}
    \label{eq:d_q}
    d_Q: Q \rightarrow B \Longleftrightarrow d_Q(q_x^t) = argmax(q_x^t)
\end{equation}

Finally, the decisions decoded would be fed to the robot simulation and turned into commands, as discussed in Subsection~\ref{ssec:robot-simulation}.

\section{Methods}
\label{sec:methods}

\begin{figure}[!ht]
    \centering
    \includegraphics[scale=0.45]{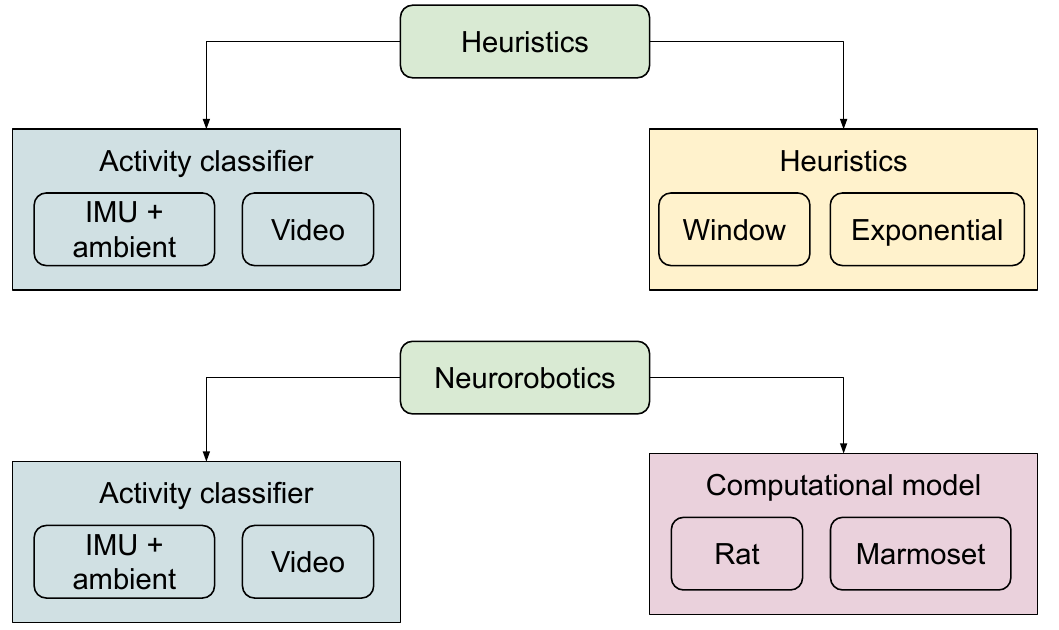}
    \caption{Factors and conditions analysed for the heuristics and neurorobotics approaches.
    For both approaches, two models for the activity recognition module were considered: the IMU + ambient and the video-based.
    For the heuristics approach, two approaches were analysed for the decision-making mechanism: window or exponential (see Subsection~\ref{ssec:robot-simulation}).
    For the neurorobotics approach, two computational models of the BG-T-C circuit were considered: the rat-based and the marmoset-based.}
    \label{fig:conditions}
\end{figure}

In Figure \ref{fig:conditions}, the different factors assessed in this work, already mentioned, are illustrated.
Both the heuristics and the neurorobotics approaches were evaluated with two different models of the activity recognition module: the IMU + ambient and the video-based (see Subsection~\ref{ssec:env-activity}).
For the heuristics approach, a couple heuristics was considered and compared: the window and the exponential (see Subsection \ref{ssec:robot-simulation}).
For the neurorobotics approach, the rat and marmoset computational models were assessed (see Subsections \ref{ssec:computational-model} and \ref{ssec:cnn-decoder}).

All code was developed in Python language.
The machine learning techniques presented for the activity recognition and the CNN-based decoder were implemented with the Tensorflow/Keras framework.
The computational models were implemented using the NetPyNE platform \cite{Dura2019}.
The robot simulation was implemented in the Gazebo simulator \cite{koenig2004design} with the Robot Operating System (ROS) \cite{quigley2009ros} as a middleware.
The next subsections will provide the implementation details of this work.

\subsection{Dataset and Classifiers}
\label{ssec:env-activity}

We have adopted the HWU-USP activities dataset \cite{Ranieri2021HumanSensors}, a multimodal and heterogeneous dataset of human activities recorded in the Robotic Assisted Living Testbed (RALT), at Heriot-Watt University (UK).
It is composed by readings of ambient sensors (e.g., switches at wardrobes and drawers, presence detectors, power measurements), inertial units attached to the waist and to the dominant wrist of the subjects, and videos.
A set of nine well-defined, pre-segmented activities of daily living was performed by the 16 participants of the data collection.
A total of $N_X=144$ recording sessions were provided, all of them pre-segmented and labelled (i.e., $X$, $A$ and $f_A$ were provided).
The length of the recording sessions varied from less than 25 to over 100 seconds, with high variance either between-classes and between-subjects.

As the activity recognition module, we have employed the framework presented and evaluated in \cite{Ranieri2021ActivitySensors}.
This was composed by different time-localised classifiers based on artificial neural networks, each focused on a particular modality (i.e., set of similar sensors) or set of modalities.
We adopted a couple pre-trained classifiers (i.e., the \textit{IMU + ambient} and the \textit{video}-based classifiers) from the framework to provide the prediction vectors, respecting the between-subjects 8-fold approach for training and evaluating.
Let those classifiers be denoted by $g_\text{I+A} \in G$ and $g_\text{video} \in G$, respectively.
Although both classifiers were described in \cite{Ranieri2021ActivitySensors}, we give a brief presentation of their architectures in the next paragraphs, for the sake of completeness.

Classifier $g_\text{I+A}$ was fed with two parallel inputs: a two-seconds-long (i.e., $100$ timesteps-long) time-window with the raw signals from the inertial sensors, and the mean values of the ambient sensors in the correspondent timestamps.
The inertial data was processed by a one-dimensional Convolutional Neural Network (CNN) \cite{Zeiler2014VisualizingNetworks}, composed of two convolutional layers interspersed with pooling layers, followed by a Long Short-Term Memory (LSTM) recurrent layer \cite{Hochreiter1997LongMemory}, generating the feature vector $v_1$.
The ambient data was processed by a single fully-connected layer, generating the feature vector $v_2$.
Both $v_1$ and $v_2$ were concatenated and sent to a softmax output layer.

Classifier $g_\text{video}$ has taken, as input, a sequence of $25$ optical flow pairs, correspondent to two seconds of video, computed with the TVL1 algorithm~\cite{Zach2007AFlow}.
The InceptionV3 CNN architecture \cite{Szegedy2016RethinkingVision} was trained to classify each optical flow pair individually.
The CNN-LSTM architecture, adopted by the authors, consisted of feeding each optical flow pair within a sequence to this pre-trained InceptionV3 module, and feeding the resulting features as inputs to each timestep of a LSTM layer, whose outputs were connected to a softmax output layer.

Both above-mentioned classifiers were endowed with softmax activation in their outputs, which ensured that the prediction vector respects a valid probability distribution.
To provide the prediction vectors, we split each recording session in $N_T=140$ timesteps, regardless to its original length, and used the referred framework to provide the predictions on each of those timesteps.
The effect is to assume that all activities have similar length, a simplification that allowed the design of more uniform and comparable experiments related to the bioinspired computational models (Subsection~\ref{ssec:computational-model}), and the robot simulation (Subsection~\ref{ssec:robot-simulation}).

The output of the activity recognition module is, for a whole recording session $x$ processed by a classifier $g$, a total of $N_X=144$ sets of prediction vectors $y_x = \{ y_x^1, \dots, y_x^{N_T} \}$, with $N_T=140$.
Outputs from both classifiers $g_\text{I+A}$ and $g_\text{video}$ were applied to all simulations, as described in the following subsections.

\subsection{Heuristics Model Implementation}
\label{ssec:heuristics-implementation}

Two policies $H$ were considered for the heuristics model, named \textit{window} or \textit{exponential}, that is, $H = \{ h_\text{window}, h_\text{exponential} \}$.
This experimental setup resulted in a total of four conditions for evaluation in the neurorobotics approach, given by the space $G \times H$.

The window policy consisted of deriving a wider prediction vector $y_w^t$, correspondent to $N_{r_w}$ timesteps.
This was done by averaging the $N_{r_w}$ most recent prediction vectors in $y_x$, from the activity recognition module, as in Equation \ref{eq:r_w}.
We have set $N_{r_w}=8$, which corresponds to windows of four seconds from the recording sessions, because this was the length of the segments considered for the neurorobotics approach (see Subsection~\ref{ssec:cnn-decoder}).
On the other hand, the exponential policy consisted of deriving a prediction vector $y_e^t$ that considered the whole sequence of previous prediction vectors in $y_x$, with an exponential decay across iterations, as in Equation \ref{eq:r_e}.
If $R=\{r_w, r_e\}$ is the set of the functions to compute $y_w^t$ and $y_e^t$, then the decision $d_r$ of the heuristics approach, for either the window or exponential policies, is given by Equation~\ref{eq:d_r}.
It is important to note that, for the window policy of the heuristics approach, as in the neurorobotics approach, the robot can only begin to move after the first four seconds of each simulation, in which it is gathering the number of prediction vectors necessary to compute the first decision.

\begin{equation}
    \label{eq:r_w}
    \begin{split}
        r_w: Y \times T \rightarrow Y \Longleftrightarrow y_w^t = r_w(y_x, t) = \frac{\sum_{i=0}^{N_{r_w}-1} y_x^{t-i}}{N_{r_w}}\\
        y_x \in Y, \quad t \in T_X | t>N_{r_w},
    \end{split}
\end{equation}
    
\begin{equation}
    \label{eq:r_e}
    \begin{split}
        r_e: Y \times T \rightarrow Y \Longleftrightarrow y_e^t =
        \begin{cases}
            r_e(y_x, t=0) = y_x^t\\
            r_e(y_x, t>0) = 0.9 \cdot r_e(y_x, t-1) + y_x^t
        \end{cases}\\
        \quad y_x \in Y, \quad t \in T_X
    \end{split}
\end{equation}

\begin{equation}
    \label{eq:d_r}
    \begin{split}
        d_R: Y \times T \times R \rightarrow B \Longleftrightarrow d_R( y_x, t, r ) = f_B( argmax[r(y_x, t)] )\\
        y_x \in Y, \quad t \in T_X, \quad r \in R
    \end{split}
\end{equation}

As a reference, we introduced an additional approach, a control condition in which the ground truth labels are directly fed to the robot simulation, providing a unique decision $d_{GT}$ every timestep, as shown by Equation~\ref{eq:d_gt}.

\begin{equation}
    \label{eq:d_gt}
    \begin{split}
        d_{GT}: X \rightarrow B \Longleftrightarrow d_{GT}(x) = f_B(f_A(x)),\\
        x \in X
    \end{split}
\end{equation}

\subsection{Neurorobotics Model Implementation}
\label{ssec:neurorobotics-configurations}

Let $M$ be the bioinspired computational model, which can be rat-based or marmoset-based, that is, $M = \{m_\text{rat}, m_\text{marmoset}\}$.
This experimental setup resulted in a total of four conditions for evaluation in the neurorobotics approach, given by the space $G \times M$.
Each independent simulation of the computational model (not to be confused with the robot simulation) was ran for each of the $N_X=144$ recording sessions under each condition being evaluated, that is, the simulations of the computational models were required to contemplate all instances in the space $X \times G \times M$.
Hence, a total of $576$ simulations of the computational model was performed.

Each of those simulations ran for $70$ seconds with sampling rate of $1,000$ Hz.
The stimuli set $S$ was updated every $0.5$ second (i.e., update frequency of 2 Hz).
This led to an adaptive dynamic that would respond to successive prediction vectors $y_x^t$, $t_\text{sim} \in \{1,\dots,N_{T_\text{sim}\}}$, with $N_{T_\text{sim}}=140$, according to the confidence of each response behaviour.
The resulting spike trains in each neuron population were converted to neural firing frequencies (for details, see \cite{Lansky2004MeanRate}), with bins of size $20$, which resulted in sequences of length $N_U=3,500$.
As stated in Subsection \ref{ssec:env-activity}, for the experiments reported in this work, $N_B=2$, hence $N_C=2$.
Considering that $N_R=8$, the multivariate time-series $u_x^{g,m} \in U$ had $N_R \cdot N_C = 16$ variables and $L_U = 3,500$ timesteps, composing a data structure with dimensions $3,500 \times 16$,

The segments for the decoder we set to $N_V = 200$ timesteps (i.e., four-seconds-long) with $75\%$ superposition (i.e., a one-second-long step between the beginning of each segment), resulting in $66$ segments.
Considering the each condition was composed of $N_x=144$ recording sessions, these simulations of the computational models led to a total of $144 \cdot 66 = 9,504$ instances $v \in V$, for each $(g,m) \in G \times M$.

The CNN architecture for decoding these time-series into response behaviours is depicted in Table~\ref{tab:cnn-decoder}.
It was composed of two convolutional layers, with $128$ and $256$ filters, respectively, interspersed with max-pooling layers.
A global average pooling operation preceded the softmax output layer, which produced the decision vector $q_x^t$.

\begin{table}[!ht]
    \centering
    \caption{Layers in the CNN-based decoder.
    The inputs to the neural network are windows of $200$ timesteps from the four cortex channels of the output signals (i.e., neural firing frequencies) of the simulations under a given condition.
    The output is a decision vector $q_x^t$ with the confidences for each response behaviour.}
    \label{tab:cnn-decoder}
    \begin{tabular}{@{}llll@{}}
    \toprule
    Layer & Type & Output shape & Free parameters \\ \midrule
    1 & Input & $200 \times 4$ & - \\
    2 & Conv1D & $200 \times 128$ & $3,712$ \\
    3 & MaxPool1D & $100 \times 128$ & - \\
    4 & Conv1D & $100 \times 256$ & $229,632$ \\
    5 & MaxPool1D & $50 \times 256$ & - \\
    6 & Global Average Pooling & $256$ & - \\
    7 & Softmax & $3$ & - \\ \bottomrule
    \end{tabular}
\end{table}

For each set of conditions, the CNN was trained in a cross-subject 8-fold cross-validation scheme, similar to the one adopted for the activity recognition module \cite{Ranieri2021ActivitySensors}.
The input data was linearly normalised to the range $[0,1]$, and the classification models were trained for $40$ epochs with batch size $32$.
The ADAM algorithm was employed, with learning rate $10^{-3}$, to optimise the categorical cross-entropy loss function.
The outputs of the evaluations were stored and organised, in order to serve as inputs to the next steps.
The resulting sequences $u_x^{g,m}$ were then introduced to the decision-making mechanism.

\subsection{Robot Behaviours}
\label{ssec:robot-simulation}

The behaviours $b \in B$ consisted of transporting an object $o \in O$, from a starting position $z \in Z$ to a fixed destination $z_\text{dest}$.
This task was adopted because it comprises a basic and generic functionality for a mobile robot in a home environment.
The associations between the behaviours and the objects are given by Equation \ref{eq:f_O}, while the associations between the objects and their starting positions in the map are given by Equation~\ref{eq:f_Z}.

\begin{align}
    \label{eq:f_O}
    f_O: B \rightarrow O \Longleftrightarrow f_O(b) = o, \quad b \in B, \quad o \in O\\
    \label{eq:f_Z}
    f_Z: O \rightarrow Z \Longleftrightarrow f_Z(o) = z, \quad o \in O, \quad z \in Z
\end{align}

At each timestep $t_\text{robot}$, a decision $d \in B \cup \{b_\emptyset\}$ (i.e., a response for each recording session $x \in X$ of the activity recognition module) was sent to the robot simulation, composed of a mobile social robot in a home environment (for details on the platforms and implementations employed, see Subsection \ref{ssec:simulation-settings}).
For the neurorobotics approach, this decision is given by Equation~\ref{eq:d_q}, already presented in Subsection~\ref{ssec:cnn-decoder}.
For the heuristics approach, the two policies mentioned (i.e., window and exponential) were evaluated.

The decisions were turned into commands to the robot following a table of rules, depicted in Table~\ref{tab:table-rules}.
A decision $d$ is sent to the robot at each timestep.
This decision can be one of the behaviours in $b \in B$ or the ''no action'' behaviour $b_\emptyset$.
Let $o_c$ be the object being carried by the robot at a certain timestep.
Two types of situations might be considered: $d \in B$ or $d=b_\emptyset$.

\begin{table}[!ht]
    \centering
    \caption{Table of rules associating a response behaviour $f(d)=b$ to an output command at each timestep $t_\text{robot}$ of the robot simulation, considering the object being carried and the current robot position.}
    \label{tab:table-rules}
    \begin{tabular}{@{}llll@{}}
    \toprule
    Decision & Object carried & Robot position & Output command \\ \midrule
    $b \in B$ & $o_c = \emptyset$ & $z \neq f_Z[f_O(b)]$ & Move towards $f_Z[f_O(b)]$ \\
    $b \in B$ & $o_c = \emptyset$ & $z = f_Z[f_O(b)]$ & Set $o_c = f_O(b)$ \\
    $b \in B$ & $o_c = f_O(b)$ & $z \neq z_\text{dest}$ & Move towards $z_\text{dest}$ \\
    $b \in B$ & $o_c = f_O(b)$ & $z = z_\text{dest}$ & Finish behaviour \\
    $b \in B$ & $o_c = o_k \in O \quad | \quad o_k \neq f_O(b)$ & $z \neq f_Z[o_k]$ & Move towards $f_Z[o_k]$ \\
    $b \in B$ & $o_c = o_k \in O \quad | \quad o_k \neq f_O(b)$ & $z = f_Z[o_k]$ & Set $o_c = \emptyset$ \\
    $b_\emptyset$ & $o_c = o_k \in O$ & $z \neq f_Z[o_k]$ & Move towards $f_Z[o_k]$ \\
    $b_\emptyset$ & $o_c = o_k \in O$ & $z = f_Z[o_k]$ & Set $o_c = \emptyset$ \\
    $b_\emptyset$ & $o_c = \emptyset$ & $\forall z \in Z$ & Wait
    \end{tabular}
\end{table}

The first type of situation is characterised by $d=b_\emptyset$, in which the robot must return any object that it may be carrying to the corresponding position, and then stand still, waiting for any further commands.
Otherwise, $d = b \in B$, the second type of situation, in which the robot is supposed to grab an object $o_c = f_O(b) \in O$ from position $f_Z[f_O(b)]$ to a destination $z_\text{dest}$.
If the robot is not carrying any object, that is, $o_c = \emptyset$, then it must move to $f_Z[f_O(b)]$ and take the object.
If it is already carrying the correct object, then it must move towards the destination $z_\text{dest}$.
If it is carrying the wrong object, it is, $o_c = o_k \in O \quad | \quad o_k \neq f_O(b)$, then it must return it to $f_Z[o_k]$.

\subsection{Robot Simulator}
\label{ssec:simulation-settings}

\begin{figure}[!ht]
    \centering
    \begin{subfigure}{0.45\textwidth}
        \centering
        \includegraphics[scale=0.6]{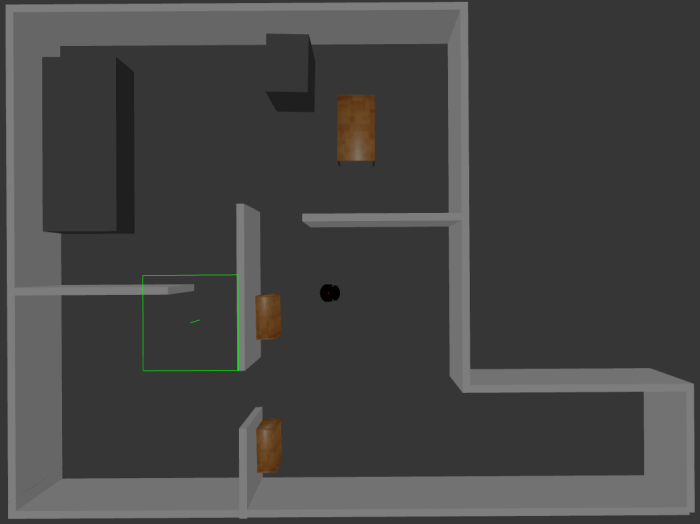}
        \caption{Map of the whole home environment employed for the experiments.}
        \label{sfig:gazebo-map}
    \end{subfigure}
    \begin{subfigure}{0.45\textwidth}
        \centering
        \includegraphics[scale=0.6]{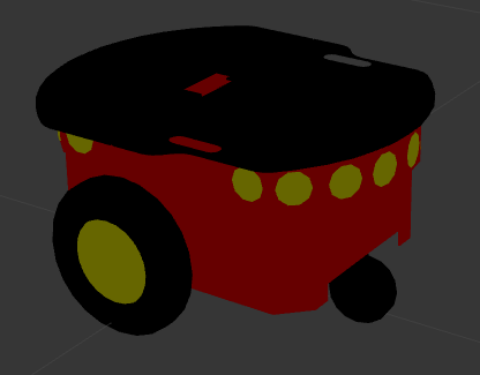}
        \caption{Simulated mobile robot.}
        \label{sfig:gazebo-robot}
    \end{subfigure}
    \begin{subfigure}{1.0\textwidth}
        \centering
        \includegraphics[scale=0.8]{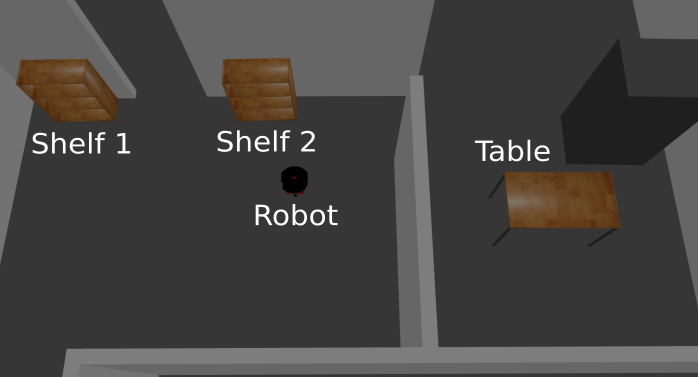}
        \caption{In a different camera angle, the section of the map in which the robot behaviours were performed, with the indications of the robot and the pieces of furniture involved in the tasks (i.e., shelf 1, shelf 2 and table).}
        \label{sfig:gazebo-furniture}
    \end{subfigure}
    \caption{Virtual environment for the robot simulation, using the Gazebo platform.}
    \label{fig:gazebo}
\end{figure}

The simulator adopted for the robotics experiments was previously made available as part of the LARa framework \cite{Ranieri2018LARa:Environments}, consisted of a robot and a software library.
The LARa robot was a mobile social robot built on the top of a Pioneer P3-DX platform, endowed with a Hokuyo laser, a mini computer, a Microsoft Kinect sensor, a microphone, a screen, and a speaker.
The LARa library was a set of functionalities implemented to control the robot based on high-level software interfaces, integrated within the Robot Operating System (ROS)~\cite{quigley2009ros}.
Besides navigation skills and a framework for human-robot interaction, this included a platform for simulation, under conditions that resembled those of the actual robot, deployed to allow offline experiments.
The Gazebo simulator \cite{koenig2004design} was employed, and a map of a typical home environment was designed, as reproduced in Figure \ref{sfig:gazebo-map}.
The simulated robot - a simplified version of the LARa robot - is shown in Figure \ref{sfig:gazebo-robot}, while the pieces of furniture employed in the experiments are shown in Figure \ref{sfig:gazebo-furniture}.

This setting comprised a realistic environment, which provided several challenging aspects resembling those of a real-world scenario, such as sensors' noise, communication delays and mechanical issues.
The ROS platform was employed to connect this simulated environment to a navigation stack, which provided a 2D occupancy grid in which each position (i.e., cell) might be considered empty, navigable or obstacle.
This representation was generated previously to the robot simulations reported here, via the GMapping algorithm \cite{grisetti2007improved} for Simultaneous Localisation and Mapping (SLAM).
The mapping algorithm ran while the robot was teleoperated through the whole environment, with the laser readings and the wheels' encoders combined to gradually compose the occupancy grid.
Once the grid was created, the Augmented Monte Carlo Localisation (AMCL) and A* algorithms could be employed as a global planner to perform autonomous navigation.
The navigation package was also endowed with a local planner, responsible for creating adaptable short-term paths for obstacle avoidance and environmental changes.

For this work, a set of two response behaviours was defined as $B = \{b_1, b_2\}$.
In Table~\ref{tab:activities}, are shown the set of daily activities from the dataset (i.e., $a_p \in A$, $p \in \{1,\dots,N_A\}$), and the expected response behaviours associated to each of those activities (i.e., $f_B(a_p)$).
These were chosen respecting semantic relationships between the activities (i.e., $b_1$ is the desired response when the user is preparing meals, and $b_2$, when he is quietly consuming or exchanging information).

\begin{table}[!ht]
\centering
\caption{List of activities provided by the HWU-USP activities dataset, and expected response behaviours in the application scenario.}
\label{tab:activities}
\begin{tabular}{@{}clc@{}}
\toprule
Activity $a$ & Description & Response behaviour $b$ \\ \midrule
$a_1$ & making a cup of tea & $b_1$ \\
$a_2$ & making a sandwich & $b_1$ \\
$a_3$ & making a bowl of cereals & $b_1$ \\
$a_4$ & using a laptop & $b_2$ \\
$a_5$ & using a phone & $b_2$ \\
$a_6$ & reading a newspaper & $b_2$ \\
$a_7$ & setting the table & $b_\emptyset$ \\
$a_8$ & cleaning the dishes & $b_\emptyset$ \\
$a_9$ & tidying the kitchen & $b_\emptyset$ \\ \bottomrule
\end{tabular}
\end{table}

These behaviours were based on the assumption that the user is located in the kitchen, and that the human activities are being monitored by sensors that are not affected by the robot actions.
The starting position for only the first robot simulation in a battery of experiments is given in Figure~\ref{fig:gazebo}. However, it had negligible effect in the overall results, since this position was not reset for each simulation, as we discuss later in this subsection.

As shown in Figure \ref{sfig:gazebo-furniture}, three pieces of furniture are considered.
These are \textit{shelf 1}, associated to the robot position $z_\text{s1} = f_Z(o_1)$, $o_1 \in O$; \textit{shelf 2}, associated to the robot position $z_\text{s2} = f_Z(o_2)$, $o_2 \in O$; and \textit{table}, the destination, associated to the robot position $z_\text{dest}$.
The two specific behaviours considered for the experiments performed, $b_1$ and $b_2$, consist, respectively, of transporting object $o_1$ from $z_1$ (i.e., shelf 1) to $z_\text{dest}$ (i.e., the table), and transporting object $o_2$ from $z_2$ (i.e., shelf 2) to $z_\text{dest}$ (i.e., the table).
Considering that \textit{shelf 2} is closer to the \textit{table} than \textit{shelf 1}, then the distances required for $b_1$ are larger than those for $b_2$.
As a consequence, it was expected that, on average, $b_1$ required more time to be completed than $b_2$.

The maximum robot simulation time was set to $N_{T_\text{robot}}=140$ seconds, with each timestep $t_\text{robot} \in \{1,\dots,N_{T_\text{robot}}\}$ corresponding to one second in the simulation.
Consequently, an expected response behaviour had to be finished within $N_{T_\text{robot}}$ seconds to be considered successfully completed.
We configured $N_{T_\text{robot}}=140$, which in exploratory experiments has shown to give a reasonable margin for the robot simulations.

A total of $N_X=144$ robot simulations was performed for each condition analysed.
The first simulation for each approach began with the robot positioned as in Figure~\ref{sfig:gazebo-furniture}.
All the next simulations began without resetting the robot position after the ending of the previous one, with only the object flag, corresponding to the object $o_c$ being carried by the robot, being cleared.
In this scenario, each simulation could be started with the robot in any of the positions in $Z$, or in locations belonging to the path between them.

\section{Results}
\label{sec:results}

Concerning the activity recognition module, its classification results are presented in \cite{Ranieri2021ActivitySensors}.
The overall accuracy registered for the classifiers were computed by taking a set of $25$ prediction vectors obtained for a recording session and averaging it.
A categorical classification was provided by returning the $argmax$ element in the averaged vector.
A cross-validation approach, following the same cross-subject partitioning adopted for evaluating the CNN-based decoder in this work, have been performed.
The accuracy reported for the modalities considered for the experiments reported here was $74.30\%$ for $g_{I+A}$, and $93.75\%$ for $g_\text{video}$.

The other modules in this work relied on important adaptations to frameworks previously implemented in related work, as happened to the computational models and the robot simulation, or to components developed from scratch, case of the CNN-based decoder.
The corresponding results are shown in the following subsections.
The classification metrics from the neural firing frequencies synthesised with the bioinspired computational models are presented in Subsection~\ref{ssec:sim-decoding}.
The outcomes of the robot simulations, in all conditions analysed, are presented in Subsection~\ref{ssec:outcomes-robotic}.

\subsection{Simulated Neural Firing Frequencies}
\label{ssec:sim-decoding}

\begin{figure}[!ht]
    \centering
    \includegraphics[scale=0.45]{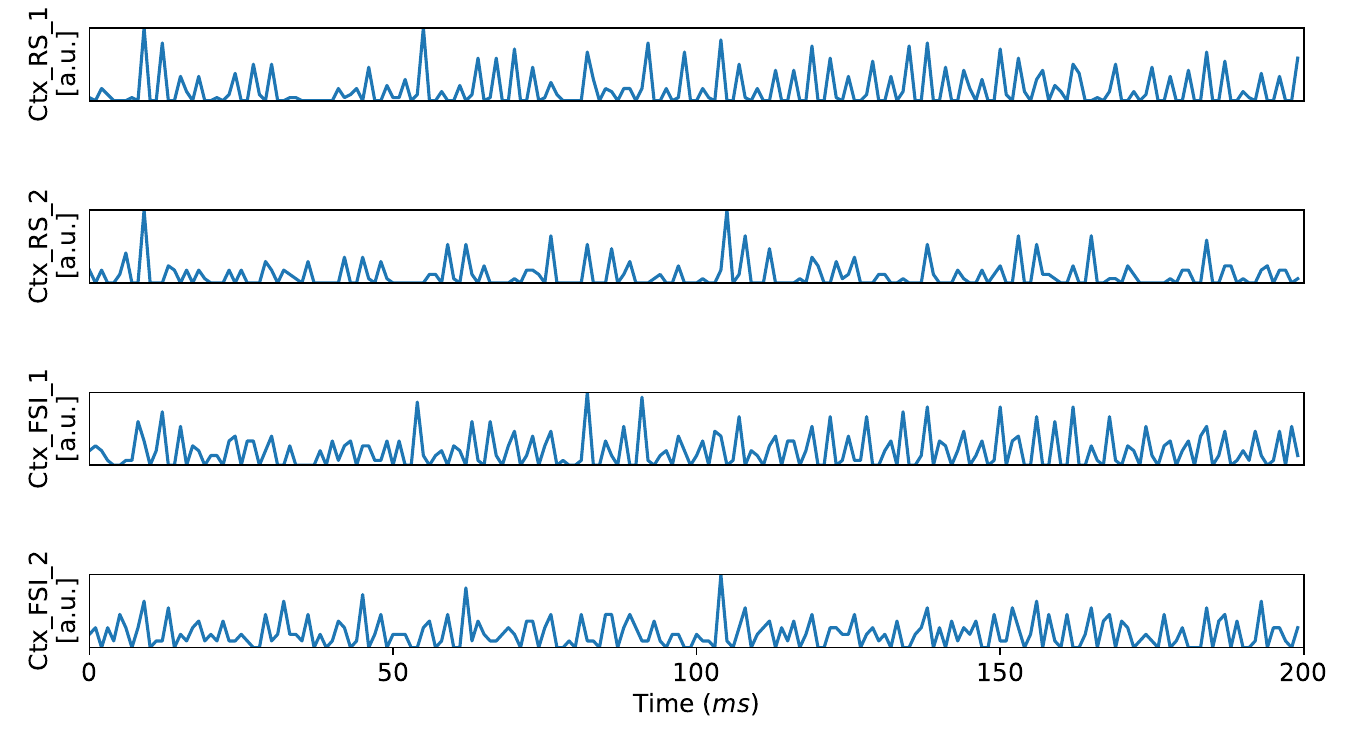}
    \caption{Sample output from the bioinspired computational model of the BG-C-T circuit.
    For the motor cortex of each channel, RS and FSI, we considered the overall mean firing rates computed with time bins of size $20$ milliseconds, and evaluated on two-seconds-long time windows.
    This data was used as input for the CNN-based decoder, in the next step of the bioinspired pipeline.}
    \label{fig:sample}
\end{figure}

A sample of the segments $v \in V$, provided in the simulations of the computational models, is shown in Figure~\ref{fig:sample}.
This was generated from a rat model, being stimulated according to an \textit{IMU + ambient} classifier as the activity recognition module.
A larger stimulus introduced to the striatum is expected to increase neural firing rates in the BG-T-C circuit, which might be propagated to the cortex.

\begin{figure}[!ht]
    \centering
    \includegraphics[scale=0.55]{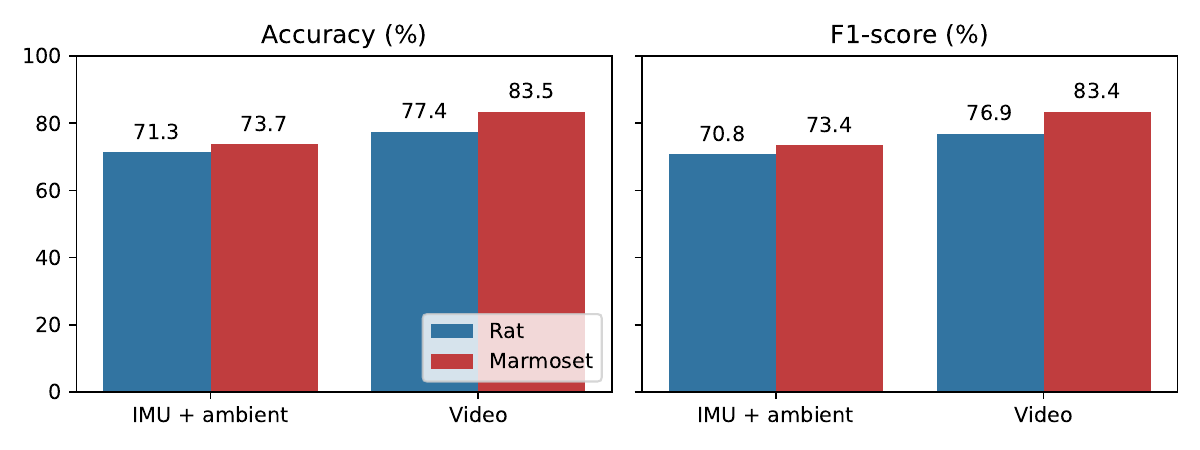}
    \caption{Accuracy and F1-score for the CNN-based decoder in classifying a MFR signal into a set of three possible decisions: B1, B2 or "no action". On choosing the models for evaluation, two factors were analysed: the modalities and models employed for activity recognition (IMU + ambient sensors or video-based)~\cite{Ranieri2021ActivitySensors}, and the computational model considered (rat-based or marmoset-based)~\cite{Kumaravelu2016ADisease,ranieri2021towardsPD}.}
    \label{fig:decoder}
\end{figure}

The overall accuracy and F1-score of the decoder, trained and evaluated according to the 8-fold cross-subject approach described, are shown in the bars plot of Figure~\ref{fig:decoder}.
The classifier used in the activity recognition module and the computational model employed are shown side-by-side.

The decoder was applied as a part of the decision-making mechanism, responsible for providing decision vectors for the robot simulation.
Hence, its results might be correlated to the correct outcomes of the decisions made during the robot simulation.
In other words, a good accuracy of the decoder might result in more correct decisions of the robot, which may more often complete the tasks with the correct outcome.
The next subsection will present the experiments performed to validate this statement.
These are the outcomes of the robot simulation not only for each of those conditions, but also for each policy employed for the heuristics approach.

\subsection{Outcomes of the Robot Simulations}
\label{ssec:outcomes-robotic}

As it was mentioned before, three possible outcomes were considered for the robot simulations, with $f_A(x)=a$ being the activity associated to a recording session $x \in X$:

\begin{itemize}
    \item \textit{Correct}, if $f_B(a) \in B$ and the activity was completed before the end of the simulation, or if $f_B(a) = b_\emptyset$ and no behaviour was completed;
    \item \textit{Incorrect}, if the robot completed a behaviour $b_\text{robot} \in B$ different from $f_B(a)$, i.e., $b_\text{robot} \neq f_B(a)$;
    \item \textit{Unfinished}, if a response behaviour $b \in B$ was expected from the robot, but no behaviour was completed before the end of the simulation.
\end{itemize}

\begin{figure}[!ht]
    \centering
    \begin{subfigure}{\textwidth}
        \centering
        \includegraphics[scale=0.55]{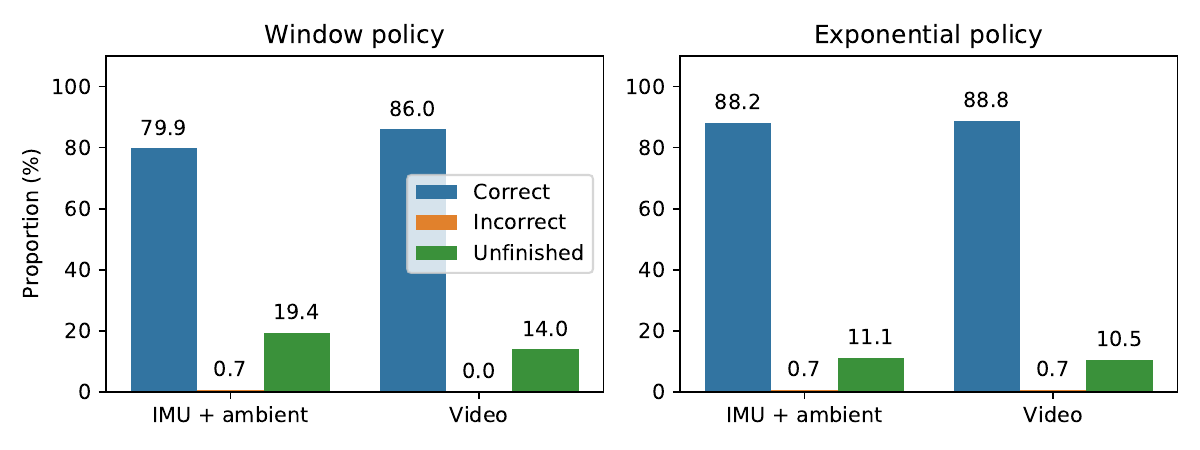}
        \caption{Outcomes for the robot simulations performed with the heuristics approach.
        Four batteries of simulations were performed, considering two factors: the classifiers employed for activity recognition (IMU + ambient sensors and video-based) and the policy for the decision-making mechanism (window or exponential) (see Figure~\ref{fig:conditions}a).}
        \label{sfig:sim-results-nonbio}
    \end{subfigure}
    \begin{subfigure}{\textwidth}
        \centering
        \includegraphics[scale=0.55]{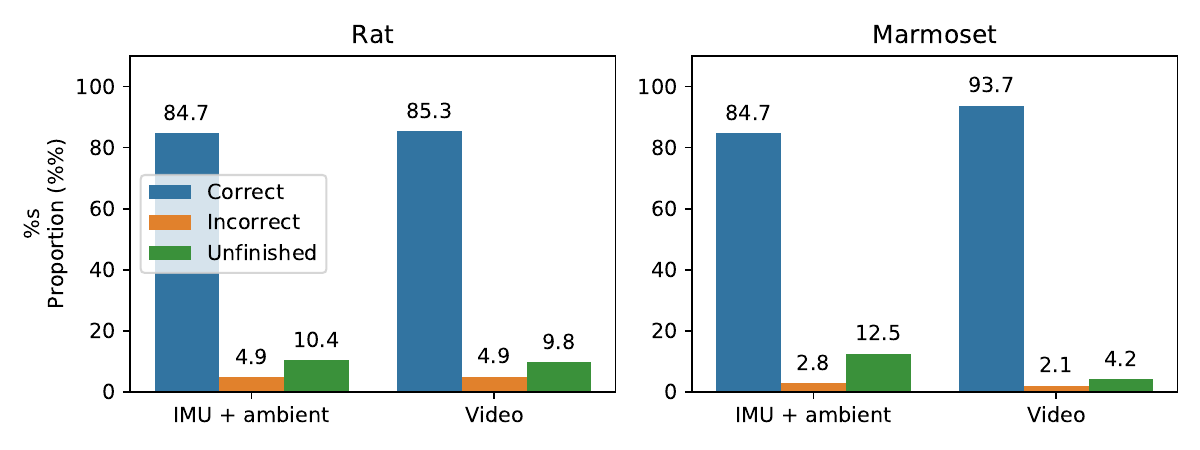}
        \caption{Outcomes for the robot simulations performed with the neurorobotics approach.
        Four batteries of simulations were performed, related to two factors analysed: the modalities and models employed for activity recognition (IMU + ambient sensors or video-based)~\cite{Ranieri2021ActivitySensors}, the computational model considered (rat-based or marmoset-based) \cite{ranieri2021towardsPD} (see Figure~\ref{fig:conditions}b).}
        \label{sfig:sim-results-bio}
    \end{subfigure}
    \caption{Outcomes for the robot simulations. Three possible outcomes were considered: the robot completed the expected (\textit{correct}) behaviour; the robot concluded the \textit{incorrect} behaviour; no behaviour was completed (\textit{unfinished}), although an action was required from the robot.}
    \label{fig:sim-results}
\end{figure}

In Subsection \ref{ssec:simulation-settings}, a control condition was introduced, with ground truth decisions being sent for the robot.
For this approach, as it was expected, all robot simulations let to the \textit{correct} outcome.
In Figure \ref{sfig:sim-results-nonbio}, the outcomes for the heuristics approach are presented, with each of the policies analysed (i.e., window and exponential) being represented in different plots, each illustrating the outcomes for each classifier considered for the activity recognition module.
The outcomes for the neurorobotics approach are shown in Figure \ref{sfig:sim-results-nonbio}, with  the classifiers for activity recognition (IMU + ambient or video) and the computational models (rat or marmoset) being represented.

\begin{figure}[!ht]
    \centering
    \includegraphics[scale=0.65]{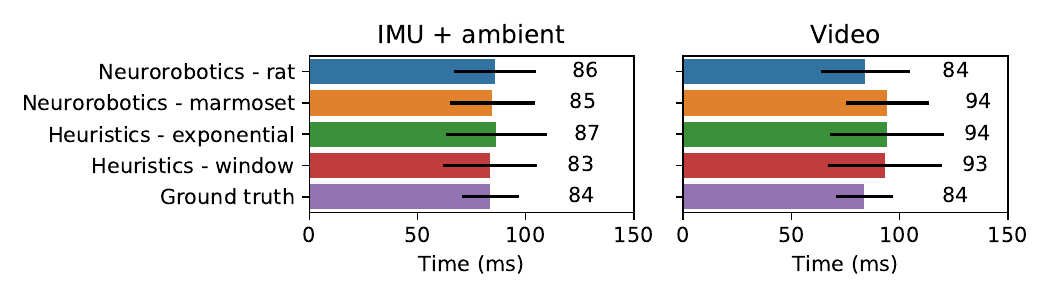}
    \caption{Average times elapsed across the 144 sequences on each simulation in which the \textit{correct} behaviour was performed. Incorrect and unfinished outcomes, as well as correct outcomes when no action was required from the robot, were not considered in this evaluation.
    All simulated models, basend on either heuristics or neurorobotics, are shown.}
    \label{fig:sim-times}
\end{figure}

The times elapsed for providing the \textit{correct} outcome, when a response behaviour was expected from the robot, were also recorded.
The mean and standard deviations, within all simulations performed for each condition, are represented in Figure \ref{fig:sim-times}.
Two separate plots were provided, separating the classifiers employed for the activity recognition module.
The ground truth approach was reproduced in both of them, since it does not depend on prediction vectors, but in the ground truth activities.

This metric considers only the outcomes completed successfully.
An approach that provides a fast response with poor accuracy would provide a low time response, though it would not necessarily provide the correct response behaviours very often.
Hence, the fact that the heuristics approach with the window policy led to a faster average response than the ground-truth condition is consistent.
Since incorrect and unfinished outcomes were not considered for the computation of this mean value, this result only shows that, for this model, the correct outcomes were mostly associated to activities that could be completed in less time (e.g., the behaviour $b_2$).

\section{Discussion}
\label{sec:discussion}

The results from the CNN-based decoder, shown in Figure \ref{fig:decoder}, confirmed some expectations regarding the output signals produced by the simulations of the computational models according to the stimuli provided: it performed better for the the video-based classifier than for the IMU + ambient, and for the marmoset-based model, compared to the rat-based.
The accuracy and F1-score metrics were very close, which considering a strictly balanced dataset, points that the results were not affected by any serious issues regarding the trade-off between precision and recall.

All evaluations led to an accuracy measure of over $70\%$ for three classes (i.e., response behaviours $b_1$, $b_2$ or $b_\emptyset$).
It is important to consider that the stimuli came from noisy prediction vectors from activity recognition algorithms, whose accuracy is variable across successive segments \cite{Ranieri2021ActivitySensors}, with overall accuracy values of $74.30\%$, for $g_{I+A}$, and $93.75\%$, for $g_\text{video}$.
These results show that the neural activity provided by the computational models could be reliably interpreted by the proposed decoder, even considering segments of limited length (i.e., four-seconds-long segments within a 70-seconds-long sequence).
Hence, this particular technique for brain signals, analysed in previous studies for processing related neuronal data of the BG-T-C circuit \cite{Oh2018ASignals,Ranieri2020UnveilingNetworks}, has shown to be suitable for the decision-making approach proposed.

Since the accuracy measure of the classifier for activity recognition was significantly higher for the video-based classifier than for the IMU + ambient, it was expected that it could be more easily decoded by the neural network, which was confirmed by the decoder results (see Figure \ref{fig:decoder}).
Also, the marmoset-based model led to better decoding performances than the rat-based model, which also meets the expectations, considering a more sophisticated morphology and dynamics in the underlying brain structures in primates than in rodents \cite{Lienard2014}.

Regarding the robot simulations, heuristics approaches were evaluated in parallel to the neurorobotics approaches.
In most experiments performed in this work, better performances were found for the models fed by the video-based classifier than those fed by the IMU + ambient classifier, which was expected, since the video classifier is expressively more accurate \cite{Ranieri2021ActivitySensors}.
As shown in Figure \ref{sfig:sim-results-nonbio}, the window policy led to a lower number of successfully completed response behaviours, especially when fed with prediction vectors coming from the IMU + ambient classifier (less accurate).
This condition may be the fairest comparison to the neurorobotics approach since it limits its decisions to data from the four-seconds-long segment that precedes a given decision, the same constraint applied to the CNN-based decoder.

In this context, the neurorobotics approach has shown to provide more accurate outcomes in most conditions, especially for the marmoset model.
For the IMU+ambient modality of the activity recogniser, the window policy of the heuristics approach led to $79.9\%$ of correct outcomes, which was surpassed by the $84.7\%$ result for either the rat or marmoset models.
For the video modality, the window policy of the heuristics approach led to $86.0\%$ of correct outcomes, which was only slightly above  the rat model, which hit $85.3\%$, and expressively below the marmoset model, which hit $93.7\%$.
These results point that the proposed neurorobotics approach, in the conditions analysed in this study, may lead to better outcomes than a simple heuristics for a real-time task of an autonomous robot.

For the exponential policy of the heuristics approach, a particularity was found: it led to similar results for either the video and IMU + ambient conditions (i.e., $88.2\%$ and $88.8\%$ of correct outcomes, respectively), both with more correct outcomes than those of the window policy.
This result is relevant, since it reveals that, by performing a long-term aggregation of prediction vectors obtained subsequently from a single recording session, it may be possible to compensate lower accuracy values provided by certain classifiers that work with different sets of sensors.
This possibility might be considered in practical applications, in which more informative modalities that usually lead to high accuracy, such as videos, may be either difficult to be obtained, due to privacy concerns \cite{FernandesJunior2016DetectionHomes}, or unfeasible to provide real-time outputs, due to the high computational cost inherent to the operations required for processing them \cite{Rodriguez-Moreno2019VideoState-of-the-Art}.

Regarding the different conditions considered for the neurorobotics approach (i.e., the activity recogniser and the computational model), the expectation was that, when applied to the robot simulation, the number of correct outcomes would be comparatively proportional to the accuracy measures of the decoder (see Figure \ref{fig:decoder}).
As shown in Figure \ref{sfig:sim-results-bio}, this expectation was met for most conditions, although some exceptions were found.

Better results for the marmoset model were expected, since the number of neurons and the connectivity are larger \cite{Prescott2006AProcessing,Koprich2017AnimalDevelopment}.
The results of the decoder, previously discussed, corroborate to this hypothesis.
For the robot simulations, considering the video modality, the marmoset model led to the best results found among all of the simulations, with $93.7\%$ of correct outcomes, against $85.3\%$ achieved by the rat model.
However, for the IMU + ambient modality, the results were similar for both models.
A possible explanation for this result is that such an increased capacity could compensate the mistakes for a more accurate activity recognise.
In other words, the prediction vectors across successive segments could assign higher confidence values (i.e., probabilities) to the expected label (i.e., the ground-truth activity) for the video-based classifier than for the IMU + ambient, and the marmoset-based model, more sophisticated, was able to take more advantage on it than the rat-based.


By measuring the time elapsed in the robot simulations with correct outcomes, we can see only modest variations across conditions.
An important observation regarding this metric is that a fast response is not necessarily an indication of a good performance, since this result is affected not only for the assertiveness of the correct outcomes (i.e., few changes of decision within a simulation), but also to the accuracy of the simulations in a given set of conditions.
For instance, a given condition may lead to fast response when it provides the correct outcome, but most simulations may lead to an incorrect or unfinished outcome.

For the neurorobotics approach, the times were approximately similar between both classifiers, except for the marmoset model, which took significantly longer to finish, on average, when fed with the video-based classifier.
Considering the heuristics approach, the video-based classifier led to clearly longer times for completing the behaviours, which was probably because some of the changes in decisions (i.e., the robot is performing behaviour $b_1$, but the decision-making mechanism changes it to $b_2$ after receiving new, updated prediction vectors) within the simulations allowed for completing more simulations with the correct outcome.
The same reason explains why the correct outcomes of the window policy for the IMU + ambient classifier led to a faster response, on average, than the ground-truth value.

\section{Conclusions and Future Work}
\label{sec:conclusion}

In this paper, we employed a neurorobotics approach based on the embodiment of validated computational models of brain structures for creating a decision-making mechanism to provide effective response behaviours to a mobile robot in a simulated environment.

The chosen application scenario was a simulated smart home where data from the sensed environment was processed with a previously designed activity recognition framework. The neurorobotics approach was compared to some heuristics. For this,  two simple heuristics were proposed and evaluated to provide real-time decisions based on the outputs from an activity recognition classifier.

The neurorobotics model used computational models (CM) of the basal ganglia-thalamus-cortex (BG-T-C) circuit, originally designed to study the underlying mechanisms of Parkinson's Disease. The CM were adapted, so that the outputs from the activity recognition module were applied as stimuli to the striatum of the circuit, and spike activity at the cortex was decoded with a convolutional neural network (CNN) to provide decisions to the robot simulation. Different conditions were analysed, including whether the computational models were based on rodent or primate models.

Results were reported with respect to the accuracy obtained for the CNN-based decoder in each condition for the computational model, and to the outcomes of the robot simulations, considering the neurorobotics and the heuristics approaches.
The expectations were met for most of the different conditions regarding the neurorobotics approaches. The primate-based computational model led to the best outcomes between the simulations analysed. 

Hence, one can conclude that the proposed neurorobotics approach is promising not only as an embedded tool for understanding the neurophysiological aspects of animal behaviour, but also as a practical component to integrate decision-making mechanisms for action selection in mobile robots engaged in human-robot-interaction scenarios.

Future work may consist of providing a real-time simulation of the proposed application scenario, with a robot placed in a physical environment in which human participants may be performing activities. This would require the integration among the different modules shown in the pipelines presented, thus ensuring that all of them can work in real-time.
Such an experiment may validate our approach in even more challenging conditions and scenarios, which may foster a wide range of applications.

\section*{Acknowledgement}
This work was funded by the Sao Paulo Research Foundation (FAPESP), grants 2017/02377-5, 2017/01687-0 and 2018/25902-0, and the Neuro4PD project - Royal Society and Newton Fund (NAF\textbackslash R2\textbackslash180773). Moioli acknowledge the support from the Brazilian institutions: INCT INCEMAQ of the CNPq/MCTI, FAPERN, CAPES, FINEP, and MEC. This research was carried out using the computational resources from the CeMEAI funded by FAPESP, grant 2013/07375-0. Additional resources were provided by the Robotics Lab within the ECR, and by the Nvidia Grants program.

This preprint has not undergone peer review or any post-submission improvements or corrections. The Version of Record of this
article is published in Cognitive Neurodynamics, and is available online at \url{https://doi.org/10.1007/s11571-022-09886-z}


%
%

\bibliographystyle{plain}      
\bibliography{references}

\end{document}